\documentclass[a4paper,fleqn]{cas-dc}

\usepackage[numbers]{natbib}

\def\tsc#1{\csdef{#1}{\textsc{\lowercase{#1}}\xspace}}
\tsc{WGM}
\tsc{QE}
\tsc{EP}
\tsc{PMS}
\tsc{BEC}
\tsc{DE}
\begin{document}
\let\WriteBookmarks\relax
\def\floatpagepagefraction{1}
\def\textpagefraction{.001}
\shorttitle{}
\shortauthors{C. Yang et~al.}

\title [mode = title]{Multi-level distortion-aware deformable network for omnidirectional image super-resolution}                      
% \tnotemark[1,2]

% \tnotetext[1]{This document is the results of the research
%    project funded by the National Science Foundation.}

% \tnotetext[2]{The second title footnote which is a longer text matters
%    to fill through the whole text width and overflow into
%    another line in the footnotes area of the first page.}

\author[1]{Cuixin Yang}
\ead{cuixin.yang@connect.polyu.hk}

\author[1]{Rongkang Dong}
\ead{rongkang97.dong@connect.polyu.hk}

\author[1]{Kin-Man Lam}
\ead{enkmlam@polyu.edu.hk}

% \credit{Conceptualization of this study, Methodology, Software}

\affiliation[1]{organization={Department of Electrical and Electronic Engineering},
                addressline={The Hong Kong Polytechnic University}, 
                city={Hong Kong},
                country={China}}

\author[2]{Yuhang Zhang}
\ead{yuhang.zhang@gzhu.edu.cn}

\affiliation[2]{organization={School of Computer Science and Cyber Engineering},
                addressline={Guangzhou University}, 
                city={Guangzhou},
                country={China}}

\author[3]{Guoping Qiu}
\ead{guoping.qiu@nottingham.ac.yk}

\affiliation[3]{organization={School of Computer Science},
                addressline={University of Nottingham}, 
                city={Nottingham},
                country={United Kingdom}}

\begin{abstract}
As augmented reality and virtual reality applications gain popularity, image processing for OmniDirectional Images (ODIs) has attracted increasing attention. 
OmniDirectional Image Super-Resolution (ODISR) is a promising technique for enhancing the visual quality of ODIs. 
Before performing super-resolution, ODIs are typically projected from a spherical surface onto a plane using EquiRectangular Projection (ERP). This projection introduces latitude-dependent geometric distortion in ERP images: distortion is minimal near the equator but becomes severe toward the poles, where image content is stretched across a wider area.
However, existing ODISR methods have limited sampling ranges and feature extraction capabilities, which hinder their ability to capture distorted patterns over large areas.
To address this issue, we propose a novel Multi-level Distortion-aware Deformable Network (MDDN) for ODISR, designed to expand the sampling range and receptive field. 
Specifically, the feature extractor in MDDN comprises three parallel branches: a deformable attention mechanism (serving as the dilation=1 path) and two dilated deformable convolutions with dilation rates of 2 and 3. This architecture expands the sampling range to include more distorted patterns across wider areas, generating dense and comprehensive features that effectively capture geometric distortions in ERP images.
The representations extracted from these deformable feature extractors are adaptively fused in a multi-level feature fusion module. Furthermore, to reduce computational cost, a low-rank decomposition strategy is applied to dilated deformable convolutions. Extensive experiments on publicly available datasets demonstrate that MDDN outperforms state-of-the-art methods, underscoring its effectiveness and superiority in ODISR.
\end{abstract}

% \includegraphics{figs/cas-grabs.pdf}

% \begin{highlights}
% \item Research highlights item 1
% \item Research highlights item 2
% \item Research highlights item 3
% \end{highlights}


\maketitle

\section{Introduction}
Omnidirectional images (ODIs), also known as $360^\circ$ images or panoramic images, serve as the foundation for augmented reality (AR) and virtual reality (VR) applications \cite{nishiyama2021360}. 
With the growing prevalence of AR/VR applications, the processing of ODIs has gained significant importance, as the quality of ODIs critically influences the immersive experience of users in AR/VR environments. ODIs are typically captured using fisheye lenses, which project scenes onto dual hemispheres. 
While high-resolution ODIs are essential for delivering realistic and detailed visualization, the prohibitive costs of optical imaging systems, coupled with storage and bandwidth constraints, often limit the resolutions of ODIs, significantly degrading the visual quality of immersive applications.

\begin{figure}
\centering
\includegraphics[width=0.49\textwidth]{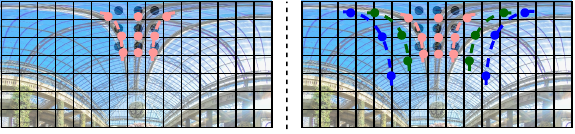}
\caption{Sampling of the previous single-level (left) and the proposed multi-level (right) distortion-aware deformable feature extractors. The centers of the regular grids represent sampling locations of a fixed, regular-shaped kernel.
Due to geometric distortion in ERP, image content, especially at high latitudes, is dramatically stretched. The sampling range of the single-level deformable feature extractor used in previous methods is limited, making it difficult to capture diverse patterns. In contrast, the proposed multi-level deformable feature extractor provides a larger sampling range, enabling better adaptation to distortion and more effective extraction of highly distorted regions.
}
\label{single_vs_multi}
\end{figure}

Image super-resolution (SR) \cite{zhang2018residual, wu2023sfhn, chi2024egovsr, wu2024real, wei2025multi} is a promising technique for solving this problem. 
Recently, researchers have focused on applying SR methods to obtain high-resolution ODIs \cite{deng2021lau, yu2023osrt, yang2024efficient, yang2025geometric}. For convenient storage and transmission, ODIs are generally projected into equirectangular projection (ERP) images because of their simplicity and compatibility with existing image processing pipelines. 
Consequently, omnidirectional image super-resolution (ODISR) is generally performed on ERP images.

In the ERP projection process, the entire spherical scene is mapped onto a 2D plane, projecting the spherical coordinates, i.e., longitude and latitude, into a rectangular grid. This method preserves the $360^\circ \times 180^\circ$ field of view (FoV) of the spherical ODI in a single, continuous flat ERP image. However, this projection introduces significant and uneven geometric distortion in ERP images, which increases with the latitude. Specifically, horizontal stretching is most extreme in the polar regions, while distortion is minimal and even trivial at the equator. 
Directly applying 2D planar image SR methods \cite{tai2017image, chen2023dual, guo2022data} to ERP images ignores geometric distortion, leading to suboptimal SR performance. Meanwhile, as illustrated in Fig.~\ref{single_vs_multi}, fixed and regular-shaped kernels, which struggle to adapt to distortion, are also unsuitable for ODISR reconstruction.

Current omnidirectional image super-resolution (ODISR) methods \cite{deng2021lau, yoon2022spheresr, yu2023osrt, yang2025geometric} consider distortion an important prior and design networks to exploit geometric distortion. 
% Given the characteristics of geometric distortion, the traditional convolutional kernels with fixed shapes struggle to adapt to the distortion for extracting a representative representation for the ERP images. 
OSRT \cite{yu2023osrt} and GDGT-OSR \cite{yang2025geometric} employ the deformable attention Transformer \cite{xia2022vision} to extract features with irregular-shaped kernels, which adapt better to distortion compared to fixed-shaped kernels. Despite the powerful geometric transformation ability of the irregular-shaped kernels, they only have a single-level sampling range, as shown in the left part of Fig.~\ref{single_vs_multi}, which makes OSRT \cite{yu2023osrt} and GDGT-OSR \cite{yang2025geometric} still struggle to adapt to the latitude-wise varying distortion. For instance, scenes at high latitudes are dramatically stretched, expanding horizontally. In these cases, the deformable extractors with only a single-level sampling range used in previous ODISR methods \cite{yu2023osrt, yang2025geometric} exhibit constrained sampling ability to capture the expanded distorted scenes. As a result, the deformable feature extractor with a single-level sampling range may only cover a few sparse distorted points around the regular kernel positions, overlooking substantial dramatically distorted contextual information essential for effective image reconstruction. 
As shown in Fig.~\ref{single_vs_multi}, the steel structure of the roof is significantly stretched, spreading across a wider area; however, the sampling range of the single-level distortion-aware deformable feature extractor remains confined to the fixed kernel, which is insufficient to capture the wider distorted area.
To address this issue, we propose the Multi-level Distortion-aware Deformable Network (MDDN), which incorporates a Multi-level Distortion-aware Deformable feature Extractor (MDDE) to achieve a larger sampling range and capture more similar distorted patterns. 
% which employs a distortion-aware deformable attention mechanism and dilated distortion-aware deformable convolutions with different dilations. 
% With multiple sampling ranges, these components extract multi-level representations for ODISR to largely adapt to the varying geometric distortion.
Fig.~\ref{single_vs_multi} illustrates that the proposed MDDE expands the sampling range to include patterns stretched across wider areas.
Specifically, the MDDE consists of a distortion-aware deformable attention mechanism and dilated distortion-aware deformable convolutions with different dilation rates. The deformable attention mechanism, with dilation equal to one, acts as the level-1 feature extractor. The dilated distortion-aware deformable convolutions with dilation rates of 2 and 3 serve as level-2 and level-3 feature extractors, respectively. Here, level-$i$ corresponds to using a dilation rate of $i$ in the Transformer or convolutional layers, which determines the sampling range of the kernels. A higher-level feature extractor corresponds to a larger sampling range, and vice versa. The level-1 feature extractor captures the basic features, while the level-2 and level-3 feature extractors complement these with features from larger sampling ranges and receptive fields. With this multi-level structure, more self-similar patterns and textures can be captured, which is beneficial for improving SR performance.
From a global-local perspective, the distortion-aware deformable attention mechanism leverages its ability to model long-range dependencies, formimg global representations for ODIs.
% Specifically,
% dilated deformable convolutions with different dilations are adopted to capture various local features, 
% we utilize the deformable attention mechanism to extract the global features, and dilated deformable convolution with different dilations are adopted to capture various local features to adaptively expand the feature extraction range of the kernels according to the distortion.
% we propose the dilated deformable convolution to adaptively expand the feature extraction range of the kernels according to the distortion. 
% The deformable convolutions with different dilations (dilation=2, 3) can extract features with wider-sampling-range dilated convolutional kernels, forming complementary and dense local representations.
The dilated deformable convolutions employ dilated convolutional kernels with different expanded sampling ranges, extracting dense and complementary local representations.
% The proposed multi-level distortion-aware network provides hierarchical representations for ODISR, including global features from the deformable attention mechanism and various local features from deformable convolutions with distinct dilations.
Furthermore, these representations cooperate adaptively in the multi-level feature fusion (MFF) module.
% we propose the mixture-of-expert mechanism to make these representations cooperate adaptively according to the latitude.
To improve the model efficiency, we adopt a low-rank decomposition strategy to decompose the weights in the deformable convolutions to reduce the computational cost. 

% avoid overfitting. With this strategy, the computational cost can be reduced.
The main contributions of this paper are summarized as follows:
\begin{itemize}
    \item We propose a novel Multi-level Distortion-aware Deformable Network (MDDN) for ODISR, which extracts features with a larger sampling range and receptive field to better adapt to distorted patterns spread across large areas.
    \item The proposed Multi-level Distortion-aware Deformable feature Extractor (MDDE) consists of a distortion-aware deformable attention mechanism and dilated distortion-aware deformable convolutions with varying dilations. These components form multi-level representations with various sampling ranges, effectively addressing the geometric distortion of ERP images by combining global with local features. The multi-level representations are adaptively fused in the proposed Multi-level Feature Fusion (MFF) module.
    % {HDDN combines global and local features in a MoE architecture, where hierarchical representations adaptively cooperate.}
    % which can enlarge the receptive field and extract multiple representative features. Specifically, we propose the dilated deformable extractor to adaptively expand the extraction range of the kernels according to the distortion. The multi-level deformable extractors dilated with different dilations can extract more related distorted contents to form more comprehensive and dense representations. 
    % \item We propose the mixture-of-expert mechanism to make these representations cooperate adaptively according to the latitude.
    \item To reduce computational cost, a low-rank decomposition strategy is employed to decompose the weights in the dilated deformable convolutions. 
    \item Extensive experimental results show that MDDN outperforms state-of-the-art (SOTA) 2D planar image SR and ODISR methods both quantitatively and qualitatively, demonstrating the effectiveness and superiority of the proposed method.
\end{itemize}

\section{Related Works}
\subsection{Image Super-resolution}
Image super-resolution \cite{qiu2000interresolution, liu2025real, huang2025lightweight} aims to reconstruct high-resolution (HR) images from low-resolution (LR) inputs, which is a classic low-level task in computer vision. Early SR methods rely on interpolation \cite{zhou2012interpolation, qiu1999progressively, gilman2008interpolation} or shallow convolutional neural networks (CNNs) \cite{dong2015image}. Since 2017, Transformers \cite{vaswani2017attention, kitaev2020reformer} have achieved excellent performances in many computer vision tasks \cite{dong2024text, gu2025optical}, including low-level computer vision tasks \cite{chen2021pre, wang2022uformer, xia2023diffir, cai2023retinexformer, tian2025degradation, he2023reti}. And Transformers have dominated the field of SR by leveraging their ability to model long-range dependencies \cite{liang2021swinir, zamir2022restormer, chen2022cross, chen2023activating, chen2023hat, conde2022swin2sr}. SwinIR \cite{liang2021swinir}, based on the hierarchical Swin Transformer \cite{liu2021swin}, employs shifted window-based self-attention to achieve computational efficiency and has become a popular architecture for image restoration. However, SwinIR's fixed-size windows limit cross-window interactions, hindering global dependency modeling. Restormer \cite{zamir2022restormer} employs a multi-scale Transformer to balance local and global feature extraction. CAT \cite{chen2022cross} introduces rectangular window self-attention, using horizontal and vertical rectangular windows to enhance global context. HAT \cite{chen2023activating, chen2023hat} combines channel attention, self-attention, and overlapping cross-attention, achieving superior perceptual quality in SR. Swin2SR \cite{conde2022swin2sr}, an evolution of SwinIR, incorporates frequency-domain attention for enhanced texture recovery.

Despite these advances, 2D planar image SR methods do not account for the varying geometric distortions present in omnidirectional images. Therefore, we introduce a novel multi-level distortion-aware deformable network tailored for ODISR.

\subsection{Omnidirectional Image Super-resolution}
Omnidirectional image super-resolution extends SR to $360^\circ$ images, which are typically represented in ERP or cubemap formats. ERP images exhibit non-uniform pixel density and geometric distortions, particularly at the poles, posing unique challenges compared to planar SR. Early ODISR methods \cite{fakour2018360, baniya2023omnidirectional} adapted CNN-based SR models either by directly fine-tuning them \cite{fakour2018360} or incorporating distortion-aware losses \cite{baniya2023omnidirectional}. However, these adaptations often struggle to address the spherical nature of omnidirectional images.

A progressive pyramid network, LAU-Net \cite{deng2021lau}, was proposed to highlight the non-uniform pixel density across latitudes. Specifically, LAU-Net hierarchically super-resolves pixels at different latitude bands using different networks. However, this approach is computationally expensive, requiring multiple levels of network training, which introduces inconsistencies across bands in the output result. To address projection-related challenges, SphereSR \cite{yoon2022spheresr} exploits Local Implicit Image Function (LIIF) \cite{chen2021learning} to obtain a continuous spherical image representation. While SphereSR flexibly handles ODIs across various projection types, it requires training separate networks for each distinct projection type. TCCL-Net \cite{chai2023tccl} introduces a Transformer and Convolution Collaborative Learning Network for ODISR, effectively modeling both short-range and long-range dependencies. The network also explores ODISR from left-right views, integrating binocular information with panoramic properties \cite{chai2023super}. DiffOSR \cite{liu2025diffosr} introduces a conditional diffusion probabilistic model and leverages geometric distortion and high-frequency sub-band from the discrete wavelet transform as prior conditions to optimize the denoising process.
BPOSR \cite{wang2024omnidirectional} utilizes two omnidirectional projections, i.e., ERP and cubemap, to facilitate interaction between these projections. OSRT \cite{yu2023osrt} employs a deformable attention mechanism to leverage distortion information. OmniSSR \cite{li2024omnissr} introduces a zero-shot method, utilizing image priors from diffusion models to achieve fidelity and realism. GDGT-OSR \cite{yang2025geometric} expands the attention area by involving multiple attention mechanisms with diverse window shapes. 

Despite these advancements, existing methods struggle with feature extraction when encountering non-uniform distorted ERP images. 
% Thus, we propose to leverage multi-level extractors to capture feature representations with hierarchical sampling ranges, allowing models to effectively adapt to distortion.
Unlike previous ODISR methods \cite{yu2023osrt, yang2025geometric}, our proposed method incorporates several feature extractors with multi-level sampling ranges to capture feature representations with hierarchical sampling ranges, which effectively capture the distorted content of ERP images and allow the model to effectively adapt to distortion. In addition, the proposed method combines local and global feature extractors, forming a more comprehensive representation for ERP images.

\subsection{Deformable Networks}
Deformable networks \cite{dai2017deformable, xia2022vision} have emerged as a powerful approach in computer vision, particularly for tasks involving geometric variations and irregular data distributions, such as ODISR. Unlike standard CNNs, which rely on fixed, grid-based receptive fields, deformable networks adaptively adjust the sampling locations of convolutional kernels or attention mechanisms based on input content. This flexibility makes them well-suited for handling non-uniform pixel density and geometric distortions inherent in ODIs, especially in ERP images. Deformable Convolutional Networks (DCN) \cite{dai2017deformable} introduce learnable offsets to standard convolutions, enabling kernels to dynamically adapt to geometric transformations in the input. EDVR \cite{wang2019edvr} handles large motions and aligns video frames using a deformable alignment module. However, due to the limitation of convolutional networks, DCN-based methods struggle to capture long-range dependencies, making them less effective for large-scale or global feature extraction.
To address this issue, Deformable Attention Transformer (DAT) \cite{xia2022vision} employs a content-adaptive attention module, which dynamically adjusts attention regions based on image features. In ODISR, deformable networks are particularly relevant due to the spherical geometry of omnidirectional images. Both OSRT \cite{yu2023osrt} and GDGT-OSR \cite{yang2025geometric} exploit deformable attention mechanisms to adapt to the inherent geometric distortions in ERP images.

In this work, to better adapt to the geometric distortion and non-uniform pixel distribution, we combine distortion-aware deformable convolutions with a distortion-aware deformable attention Transformer to thoroughly exploit local and global features, thereby improving the performance of ODISR.

\begin{figure}
\centering
\includegraphics[width=0.49\textwidth]{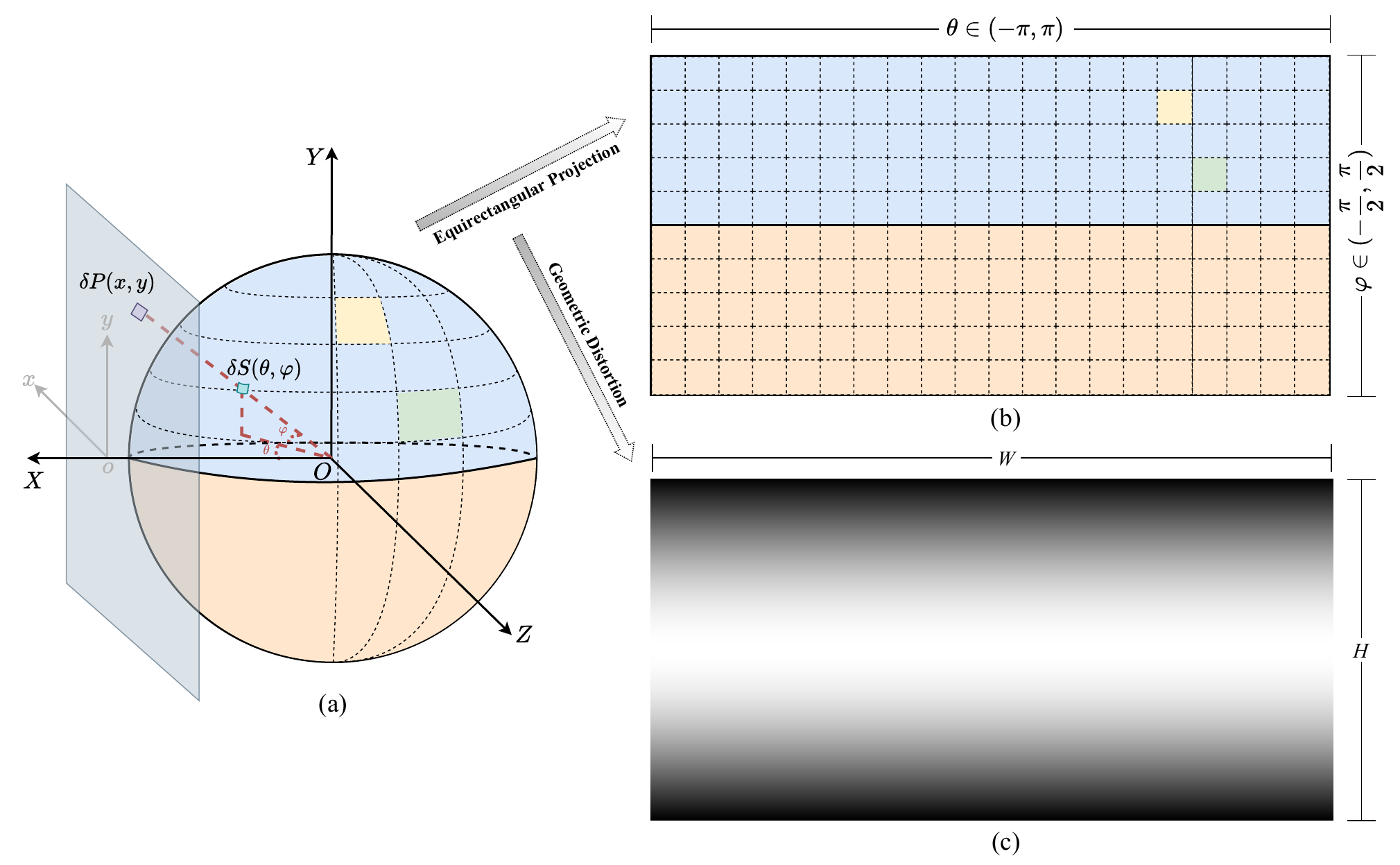}
\caption{(a) Relationship between the sphere and the projection plane. (b) Visualization of equirectangular projection. The omnidirectional images are mapped onto a 2D plane through equirectangular projection. (c) Distortion map. Darker areas indicate greater distortion, whereas lighter areas show less distortion. The equirectangular projection causes geometric distortion in the equirectangular projection images. The distortion intensifies with rising latitude. With the Equator as the symmetry axis, the geometric distortion in the Northern and Southern Hemispheres is symmetric.}
\label{projection}
\end{figure}

\begin{figure*}[!t]
\centering
\includegraphics[width=0.9\textwidth]{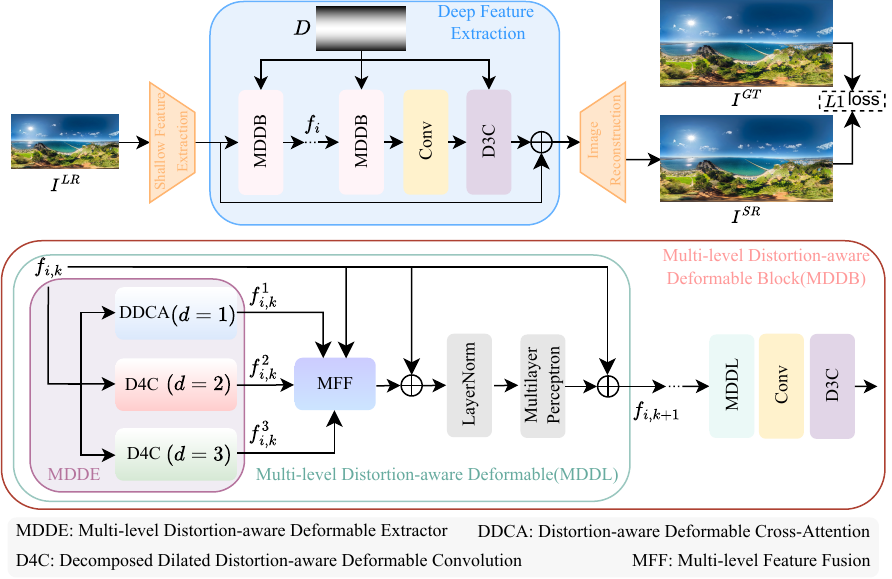}
\caption{Overview of the architecture of the proposed network. $D$ denotes the distortion map. $d$ represents dilation.}
\label{overview}
\end{figure*}
\section{Methodology}
\subsection{Fundamentals of Distortion in ODIs}
Fig.~\ref{projection}(a) depicts the geometric transformation between the sphere and the projection plane. $d\theta d\varphi$ represents a microunit whose area is $\delta S(\theta, \varphi)$ on the sphere centered at $(\theta, \varphi)$, and $dxdy$ is its corresponding counterpart whose area is $\delta P(x, y)$ on the projection plane centered at $(x, y)$.
Each point $(\theta, \varphi)$ on the spherical surface corresponds to a specific point $(x, y)$ on the projection plane, which can be illustrated as follows: 
\begin{equation}
\begin{aligned}
        & x = f(\theta, \varphi) \\
        & y = g(\theta, \varphi),
\end{aligned}
\end{equation}
where $\theta \in(-\pi, \pi)$ and $\varphi \in(-\frac{\pi}{2}, \frac{\pi}{2})$. With this relationship, the geometric transformation between $dxdy$ and $d\theta d\varphi$ can be achieved by the Jacobian matrix. Specifically, the Jacobian determinant $J(\theta, \varphi)$ is defined as follows:
\begin{equation}
    J(\theta, \varphi)=\frac{\partial (x, y)}{\partial (\theta, \varphi)}=
\begin{vmatrix}
\frac{\partial x}{\partial \theta} & \frac{\partial x}{\partial \varphi} \\
\frac{\partial y}{\partial \theta} & \frac{\partial y}{\partial \varphi} \\
\end{vmatrix}.
\label{jacobian}
\end{equation}
The area of the microunit $d\theta d\varphi$ on the sphere, i.e., $\delta S(\theta, \varphi)$, can be expressed as $\cos(\varphi)|d\theta d\varphi|$, and the area of the microunit $dxdy$ on the projection plane, i.e., $\delta P(x, y)$, is equal to $|dxdy|$.
The stretching ratio $\rho$ between the spherical surface and the projection plane is defined as follows:
\begin{equation}
    \rho(x, y)=\frac{\delta S(\theta, \varphi)}{\delta P(x, y)}=\frac{\cos(\varphi)|d\theta d\varphi|}{|dxdy|}=\frac{\cos(\varphi)}{|J(\theta, \varphi)|}.
\end{equation}

For ERP shown in Fig.~\ref{projection}(b), the relationship between $(x, y)$ and $(\theta, \varphi)$ is as follows:
\begin{equation}
    \begin{aligned}
            & x = \theta \\
            & y = \varphi.
    \end{aligned}
\end{equation}
Then, the Jacobian determinant for ERP $J_{ERP}(\theta, \varphi)$ is equal to 1. Thus, the stretching ratio for ERP $\rho_{ERP}$ is simplified as follows:
\begin{equation}
    \rho_{ERP}(x, y)=\frac{\cos(\varphi)}{|J_{ERP}(\theta, \varphi)|}=\cos(\varphi)=\cos(y),
\end{equation}
where $y\in(-\frac{\pi}{2}, \frac{\pi}{2})$.

The projection leads to geometric distortion in the ERP images. 
For an LR image $I\in\mathbb{R}^{C\times H\times W}$, its distortion map $D \in\mathbb{R}^{1\times{H}\times{W}}$, as shown in Fig.~\ref{projection}(c), is defined as follows:
\begin{equation}
    D=\cos\frac{(h+0.5-H/2)\pi}{H},
\end{equation}
where $h$ denotes the current height of the point position. It can be inferred that the distortion increases with increasing latitude and is symmetric in the two hemispheres, varying along the latitude.

\subsection{Overview of the Framework}
The overview of the architecture of the proposed MDDN is shown in Fig.~\ref{overview}. Our proposed network is composed of three main components, for shallow feature extraction, deep feature extraction, and image reconstruction. The component for deep feature extraction consists of several multi-level distortion-aware deformable blocks (MDDBs), a convolutional layer, and a decomposed distortion-aware deformable convolution (D3C). The bottom part of Fig.~\ref{overview} shows the structure of MDDB, which comprises several multi-level distortion-aware deformable layers (MDDLs), a convolutional layer, and a D3C layer. To capture dense and comprehensive representations that effectively model distorted patterns, each MDDL takes the extracted features $f_{i,k}$ (where $i$ denotes the $i$-th MDDB and $k$ denotes the $k$-th MDDL) and feeds them into the multi-level distortion-aware deformable extractor (MDDE). The MDDE consists of two decomposed dilated distortion-aware deformable convolutions (D4Cs) with dilations of 2 and 3, and a distortion-aware deformable cross-attention (DDCA) mechanism. To ensure better cooperation among these three branches, we propose the Multi-level Feature Fusion (MFF) module, which adaptively fuses the corresponding output features. $L1$ loss is the training loss, which is used to calculate the pixel-wise differences between the final SR result $I^{SR}$ and its corresponding ground-truth $I^{GT}$.

\subsection{Multi-level Distortion-aware Deformable Extractor}
To densely capture similar patterns and adapt to distorted content, we propose multi-level distortion-aware deformable extractor (MDDE) whose kernels are of different sampling ranges.
% These extractors consist of dilated distortion-aware deformable convolutions and a deformable attention mechanism. 
Specifically, each MDDE includes two decomposed dilated distortion-aware deformable convolutions (D4Cs) with different dilations ($d=2$ and $d=3$), and one distortion-aware deformable cross-attention (DDCA) module. The DDCA module can be considered as level-1 extractor ($d=1$), while the D4Cs serve as level-2 ($d=2$) and level-3 ($d=3$) extractors.
By leveraging multi-level extractors, MDDE effectively adapts to geometric distortions with various kernels of multi-level sampling ranges, contributing to a broader receptive field. This allows more similar and related features to be captured, resulting in improved reconstruction of the current patch. 

\subsubsection{Distortion-aware Deformable Cross-Attention}
\begin{figure*}
\centering
\includegraphics[width=\textwidth]{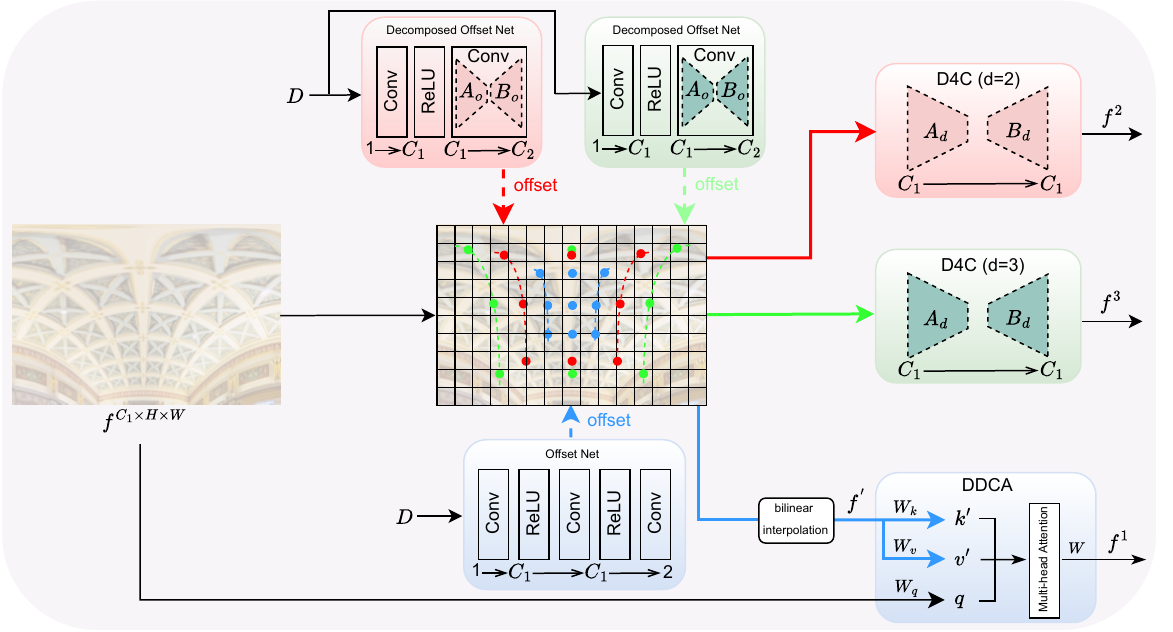}
\caption{Structure of Multi-level Distortion-aware Deformable Extractor (MDDE). It consists of a distortion-aware deformable cross-attention (DDCA) mechanism and two decomposed dilated distortion-aware deformable convolution (D4C) layers with $d=2$ and $d=3$, and their offset networks. The numbers of input and output channels are shown below the layers.}
\label{MLDE}
\end{figure*}

The DDCA mechanism, as the level-1 feature extractor, plays a critical role in the proposed MDDE due to its strong transformation ability (refer to Sec.~\ref{abla_mdde}).
The DDCA mechanism and its related offset network are shown in the blue box of Fig.~\ref{MLDE}. The distortion map $D$ is first fed into the offset network, which generates the corresponding offsets. The feature $f$ is then warped according to these offsets through bilinear interpolation. The warped feature $f^{'}$ is projected to form the key and value features, while the original feature $f$ is projected as the query feature. This process can be formulated as follows: 
\begin{equation}
    \Delta{p^1}=\mathcal{N}^1(D), 
    f^{'}=\mathcal{BI}(f, p^1 + \Delta{p^1}),
\end{equation}
\begin{equation}
    f^{1}=\mathcal{CA}(W_{q}f, W_{k}f^{'}, W_{v}f^{'}),
\end{equation}
where $\mathcal{N}^{1}$ denotes the offset network for DDCA. Following OSRT \cite{yu2023osrt}, the offset network consists of several convolutional layers and activation layers with the numbers of input and output channels shown below them. $\mathcal{BI}$ denotes the bilinear interpolation operation, and $\mathcal{CA}$ denotes the cross-attention mechanism. $p^1$ represents the locations on the regular grid, and $\Delta{p^1}$ represents the corresponding generated offsets. The warped feature $f^{'}$ and the original feature $f$ interact in the cross-attention mechanism, enabling DDCA to generate a global representation for the inputs.

Although DDCA can effectively transform distorted features, relying on sampling kernels with only a single-level sampling range limits its ability to accurately model features in heavily distorted regions. To address this limitation, we introduce D4C layers, which enhance the multi-level feature representations of the network by extracting local distorted contents using dilated convolutional kernels with wider sampling ranges, complementing the global representation from DDCA.

\subsubsection{Decomposed Dilated Distortion-aware Deformable Convolution}
Wide-range repetitive and similar patterns are often observed in large FoV images, such as the ground. Due to geometric distortion, these patterns are further expanded in ERP images. In addition, the levels of distortion vary along the latitudes. Therefore, the sampling ability of the single-level feature extractor is limited to capturing similar patterns for ERP images.
To better adapt feature representations to the geometric distortion in ERP images, we propose integrating DDCA with D4C layers of varying dilations. As shown in Fig.~\ref{MLDE}, the red and green boxes depict the D4C layers with $d=2$ and $d=3$, respectively, and their corresponding offset networks. By using different dilations, the dilated convolutional kernels expand their sampling ranges to different degrees as shown in Fig.~\ref{MLDE}, thus forming distinct local representations for the inputs, which can complement the global representation from DDCA. Furthermore, based on geometric distortion, the offset networks generate adaptive offsets to adjust the sampling locations of the kernel, adapting to the varying distortion. This process can be expressed as follows:
\begin{equation}
    \Delta{p^2}=ONet2(D), 
    f^2=D4C^2(f,\Delta{p^2}),
\end{equation}
\begin{equation}
    \Delta{p^3}=ONet3(D), 
    f^3=D4C^3(f,\Delta{p^3}),    
\end{equation}
where $ONet2$ denotes the offset network for the D4C layer with $d=2$, which is denoted as \textit{$D4C^2$}, i.e., level-2 extractor. Similarly, $ONet3$ denotes the offset network for the D4C layer with $d=3$, which is denoted as \textit{$D4C^3$}, i.e., level-3 extractor. Additionally, $\Delta{p^2}$ and $\Delta{p^3}$ represent the generated offsets for the level-2 and level-3 extractors, respectively.

As shown in Fig.~\ref{MLDE}, $ONet2$ and $ONet3$ share the same structure, consisting of two convolutional layers and an activation layer. The numbers of input and output channels are shown below the corresponding layers. Specifically, the distortion map $D\in\mathbb{R}^{1\times{H}\times{W}}$ has only one channel, therefore, the first convolutional layer has one input channel and $C_{1}$ output channels. The second convolutional layer has $C_{1}$ input channels and $C_{2}$ output channels, where $C_{1} \gg 1$ and $C_{2} \gg 1$. The weight matrix of the second convolutional layer is $W\in\mathbb{R}^{{C_1}\times{C_2}\times{N}\times{N}}$, where $N$ represents the kernel size. Considering the substantial number of training parameters in the second convolutional layer, we adopt a low-rank decomposition strategy to decompose its weight into two low-rank matrices, reducing the parameter count. Details of the decomposition process will be introduced in the following section.

Overall, in MDDE, the DDCA and D4C layers provide features extracted by various kernels with multi-level sampling ranges, leading to better adaptation of the latitude-wise varying geometric distortion and a more comprehensive representation of distorted ERP images.

\subsubsection{Low-rank Decomposition}
To reduce the number of training parameters,
% efficiently adapt the proposed network to the ODISR task, 
we adopt low-rank decomposition not only in offset networks but also in distortion-aware deformable convolutions, including D3Cs and D4Cs.
We mainly decompose a large and highly complex weight into two low-rank matrices. Specifically, the convolutional weight $W\in \mathbb{R}^{{C_i}\times{C_o}\times{N}\times{N}}$, where $C_i$ and $C_o$ represent the number of input and output channels, and $N$ denotes the kernel size, is first reshaped into a 2D weight matrix $W^{'}\in\mathbb{R}^{{C_i\times{C_oN^2}}}$. It is then decomposed into two low-rank weight matrices as follows:
% \begin{center}
\begin{equation}
    W^{'}=AB^T,
\end{equation}
% \end{center}
where $A\in\mathbb{R}^{C_i\times r}$ and $B\in\mathbb{R}^{C_oN^2\times r}$, with the rank $r\ll \textit{min}(C_i, C_o)$. $W^{'}$ is then reshaped back into its original shape and used to convolve the features. In Fig.~\ref{MLDE}, $A_o$ and $B_o$ represent the decomposed low-rank weight matrices of the decomposed offset networks, while $A_d$ and $B_d$ represent the decomposed low-rank weight matrices of the D4Cs.
By applying low-rank decomposition, a large number of training parameters can be reduced, leading to significant savings in memory footprint and computational cost.

\subsection{Multi-level Feature Fusion}
\begin{figure}
\centering
\includegraphics[width=0.48\textwidth]{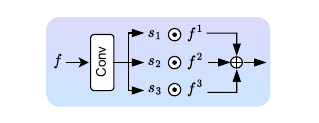}
\caption{Multi-level Feature Fusion.}
\label{MoE}
\end{figure}
In ERP images, distortion varies across different regions. Specifically, scenes at higher latitudes experience more extreme stretching, which require a larger sampling range, while distortion is reduced at lower latitudes, where a moderate sampling range is needed. 
% Due to this geometric distortion, various areas should be treated differently. 
This geometric distortion necessitates distinct treatment for areas with different latitudes.
To effectively adapt to ERP's geometric characteristics, we adaptively fuse multi-level representations in the multi-level feature fusion (MFF) module. Specifically, pixel-wise spatial attention is generated to adjust the contribution of each feature extractor.
As shown in Fig.~\ref{MoE}, the features $f$ are first convolved to generate spatial attentions for the three feature extractors. Then, the spatial attentions are split into three groups, denoted as $s_1$, $s_2$, and $s_3$, which multiply with the outputs from the three feature extractors. For each pixel in the feature maps, distinct feature extractors contribute differently.
Assigning different weights to the outputs from different extractors can effectively adapt to the variation of the distortion across the whole image. Subsequently, the three weighted outputs are aggregated through pixel-wise summation to obtain the fused features.

\section{Experiment}
\begin{table*}[]
\centering
\caption{Quantitative comparison of different methods on ODI-SR \cite{deng2021lau}, SUN 360 Panorama \cite{xiao2012recognizing} and Flickr360 val \cite{cao2023ntire} datasets. The upscaling factors are 4, 8 and 16. The best results are highlighted in bold.}
% \resizebox*{18cm}{0.5\textheight}{
\resizebox{\textwidth}{!}{
\begin{tabular}{c|cccccccccccc}
\hline
         & \multicolumn{12}{c}{ODI-SR \cite{deng2021lau}}                                                                                                                                                                                                                                \\ \cline{2-13} 
         & \multicolumn{4}{c|}{x4}                                                                  & \multicolumn{4}{c|}{x8}                                                                  & \multicolumn{4}{c}{x16}                                             \\ \cline{2-13} 
         & PSNR           & SSIM            & WS-PSNR        & \multicolumn{1}{c|}{WS-SSIM}         & PSNR           & SSIM            & WS-PSNR        & \multicolumn{1}{c|}{WS-SSIM}         & PSNR           & SSIM            & WS-PSNR        & WS-SSIM         \\ \hline\hline
Bicubic  & 25.59          & 0.7118          & 24.95          & \multicolumn{1}{c|}{0.6923}          & 23.75          & 0.6442          & 23.08          & \multicolumn{1}{c|}{0.6123}          & 22.13          & 0.6116          & 21.44          & 0.5732          \\
SwinIR \cite{liang2021swinir}   & 27.21          & 0.7683          & 26.53          & \multicolumn{1}{c|}{0.7542}          & 25.12          & 0.6873          & 24.41          & \multicolumn{1}{c|}{0.6616}          & 23.35          & 0.6395          & 22.60          & 0.6052          \\
HAT \cite{chen2023activating, chen2023hat}      & 27.29          & 0.7711          & 26.60          & \multicolumn{1}{c|}{0.7567}          & 25.01          & 0.6791          & 24.27          & \multicolumn{1}{c|}{0.6509}          & 23.34          & 0.6354          & 22.58          & 0.5997          \\
RGT \cite{chen2024recursive}     & 27.25          & 0.7708          & 26.55          & \multicolumn{1}{c|}{0.7566}          & 25.14          & 0.6876          & 24.42          & \multicolumn{1}{c|}{0.6621}          & 23.32          & 0.6380          & 22.57          & 0.6037          \\ \hline
OSRT \cite{yu2023osrt}    & 27.28          & 0.7711          & 26.58          & \multicolumn{1}{c|}{0.7559}          & 25.20          & 0.6901          & 24.46          & \multicolumn{1}{c|}{0.6635}          & 23.46          & 0.6410          & 22.68          & 0.6063          \\
BPOSR \cite{wang2024omnidirectional}   & 26.10          & 0.7231          & 25.43          & \multicolumn{1}{c|}{0.7038}          & 24.15          & 0.6524          & 23.42          & \multicolumn{1}{c|}{0.6213}          & 22.54          & 0.6168          & 21.77          & 0.5791          \\
GDGT-OSR \cite{yang2025geometric} & 27.29          & 0.7706          & 26.59          & \multicolumn{1}{c|}{0.7557}          & 25.22          & 0.6901          & 24.48          & \multicolumn{1}{c|}{0.6636}          & 23.45          & 0.6408          & 22.68          & 0.6063          \\
MDDN(Ours)     & \textbf{27.39} & \textbf{0.7751} & \textbf{26.68} & \multicolumn{1}{c|}{\textbf{0.7595}} & \textbf{25.27} & \textbf{0.6926} & \textbf{24.52} & \multicolumn{1}{c|}{\textbf{0.6659}} & \textbf{23.52} & \textbf{0.6423} & \textbf{22.73} & \textbf{0.6077} \\ \hline\hline
         & \multicolumn{12}{c}{SUN 360 Panorama \cite{xiao2012recognizing}}                                                                                                                                                                                                                     \\ \cline{2-13} 
         & \multicolumn{4}{c|}{x4}                                                                  & \multicolumn{4}{c|}{x8}                                                                  & \multicolumn{4}{c}{x16}                                             \\ \cline{2-13} 
         & PSNR           & SSIM            & WS-PSNR        & \multicolumn{1}{c|}{WS-SSIM}         & PSNR           & SSIM            & WS-PSNR        & \multicolumn{1}{c|}{WS-SSIM}         & PSNR           & SSIM            & WS-PSNR        & WS-SSIM         \\ \hline\hline
Bicubic  & 25.29          & 0.6993          & 24.90          & \multicolumn{1}{c|}{0.7083}          & 23.23          & 0.6290          & 22.72          & \multicolumn{1}{c|}{0.6241}          & 21.52          & 0.5978          & 20.92          & 0.5850          \\
SwinIR \cite{liang2021swinir}  & 27.54          & 0.7739          & 27.47          & \multicolumn{1}{c|}{0.7933}          & 25.01          & 0.6879          & 24.61          & \multicolumn{1}{c|}{0.6947}          & 22.87          & 0.6288          & 22.26          & 0.6223          \\
HAT \cite{chen2023activating, chen2023hat}     & 27.67          & 0.7777          & 27.60          & \multicolumn{1}{c|}{0.7970}          & 24.80          & 0.6743          & 24.34          & \multicolumn{1}{c|}{0.6765}          & 22.80          & 0.6237          & 22.21          & 0.6154          \\
RGT \cite{chen2024recursive}     & 27.61          & 0.7770          & 27.54          & \multicolumn{1}{c|}{0.7965}          & 25.04          & 0.6886          & 24.64          & \multicolumn{1}{c|}{0.6955}          & 22.82          & 0.6276          & 22.24          & 0.6210          \\ \hline
OSRT \cite{yu2023osrt}    & 27.65          & 0.7773          & 27.55          & \multicolumn{1}{c|}{0.7956}          & 25.16          & 0.6919          & 24.70          & \multicolumn{1}{c|}{0.6978}          & 23.04          & 0.6316          & 22.40          & 0.6253          \\
BPOSR \cite{wang2024omnidirectional}   & 25.99          & 0.7201          & 25.71          & \multicolumn{1}{c|}{0.7329}          & 23.67          & 0.6429          & 23.15          & \multicolumn{1}{c|}{0.6403}          & 21.94          & 0.6050          & 21.30          & 0.5934          \\
GDGT-OSR \cite{yang2025geometric} & 27.68          & 0.7770          & 27.58          & \multicolumn{1}{c|}{0.7956}          & 25.19          & 0.6920          & 24.74          & \multicolumn{1}{c|}{0.6982}          & 23.04          & 0.6316          & 22.41          & 0.6252          \\
MDDN(Ours)     & \textbf{27.81} & \textbf{0.7823} & \textbf{27.71} & \multicolumn{1}{c|}{\textbf{0.8002}} & \textbf{25.27} & \textbf{0.6952} & \textbf{24.80} & \multicolumn{1}{c|}{\textbf{0.7012}} & \textbf{23.12} & \textbf{0.6336} & \textbf{22.48} & \textbf{0.6274} \\ \hline\hline
         & \multicolumn{12}{c}{Flickr360 val \cite{cao2023ntire}}                                                                                                                                                                                                                        \\ \cline{2-13} 
         & \multicolumn{4}{c|}{x4}                                                                  & \multicolumn{4}{c|}{x8}                                                                  & \multicolumn{4}{c}{x16}                                             \\ \cline{2-13} 
         & PSNR           & SSIM            & WS-PSNR        & \multicolumn{1}{c|}{WS-SSIM}         & PSNR           & SSIM            & WS-PSNR        & \multicolumn{1}{c|}{WS-SSIM}         & PSNR           & SSIM            & WS-PSNR        & WS-SSIM         \\ \hline\hline
Bicubic  & 27.26          & 0.7579          & 26.90          & \multicolumn{1}{c|}{0.7496}          & 24.59          & 0.6697          & 24.22          & \multicolumn{1}{c|}{0.6497}          & 22.45          & 0.6280          & 22.11          & 0.6022          \\
SwinIR \cite{liang2021swinir}  & 30.11          & 0.8300          & 29.62          & \multicolumn{1}{c|}{0.8235}          & 26.79          & 0.7265          & 26.25          & \multicolumn{1}{c|}{0.7088}          & 24.23          & 0.6618          & 23.69          & 0.6370          \\
HAT \cite{chen2023activating, chen2023hat}     & 30.27          & 0.8333          & 29.76          & \multicolumn{1}{c|}{0.8264}          & 26.57          & 0.7165          & 26.00          & \multicolumn{1}{c|}{0.6960}          & 24.18          & 0.6577          & 23.63          & 0.6318          \\
RGT \cite{chen2024recursive}     & 30.18          & 0.8326          & 29.67          & \multicolumn{1}{c|}{0.8260}          & 26.79          & 0.7265          & 26.25          & \multicolumn{1}{c|}{0.7089}          & 24.17          & 0.6600          & 23.66          & 0.6355          \\ \hline
OSRT \cite{yu2023osrt}    & 30.24          & 0.8331          & 29.70          & \multicolumn{1}{c|}{0.8253}          & 26.94          & 0.7305          & 26.35          & \multicolumn{1}{c|}{0.7113}          & 24.40          & 0.6641          & 23.81          & 0.6385          \\
BPOSR \cite{wang2024omnidirectional}   & 28.17          & 0.7746          & 27.70          & \multicolumn{1}{c|}{0.7644}          & 25.26          & 0.6812          & 24.74          & \multicolumn{1}{c|}{0.6597}          & 23.09          & 0.6355          & 22.60          & 0.6094          \\
GDGT-OSR \cite{yang2025geometric} & 30.24          & 0.8323          & 29.71          & \multicolumn{1}{c|}{0.8249}          & 26.94          & 0.7300          & 26.36          & \multicolumn{1}{c|}{0.7111}          & 24.41          & 0.6640          & 23.83          & 0.6385          \\
MDDN(Ours)     & \textbf{30.44} & \textbf{0.8380} & \textbf{29.87} & \multicolumn{1}{c|}{\textbf{0.8295}} & \textbf{27.05} & \textbf{0.7339} & \textbf{26.45} & \multicolumn{1}{c|}{\textbf{0.7144}} & \textbf{24.48} & \textbf{0.6662} & \textbf{23.88} & \textbf{0.6403} \\ \hline
\end{tabular}}
\label{SOTA}
\end{table*}

\subsection{Experimental Settings}
\subsubsection{Datasets}
In our experiments, we use the ODI-SR \cite{deng2021lau} and Flickr360 \cite{cao2023ntire} datasets for training. The ODI-SR dataset and the Flickr360 dataset contain 1,200 and 3,000 HR training images, respectively. The test datasets include the ODI-SR test dataset \cite{deng2021lau}, SUN360 Panorama dataset \cite{xiao2012recognizing}, and Flick360-val dataset \cite{cao2023ntire}. These three test datasets contain 100, 100, and 50 HR images, respectively. The resolution of the HR images is $1024\times2048$ pixels. Following OSRT \cite{yu2023osrt}, we employ fisheye downsampling to obtain HR-LR image pairs. We conduct experiments using $\times4$, $\times8$, and $\times16$ downsampling factors.
\subsubsection{Implementation Details}
During the training process, the batch size is set to 4, with a patch size of $256\times256$ pixels. The framework is trained for 500,000 iterations with an initial learning rate of $2\times 10^{-4}$, which is halved at the $250,000^{th}$, $400,000^{th}$, $450,000^{th}$, and $475,000^{th}$ iterations. $L1$ loss is adopted as the training loss, which is used to calculate the mean absolute error between each pixel in the output SR result and the corresponding ground truth. The Adam optimizer \cite{kingma2015adam} is used, with parameters $\beta_1=0.9$ and $\beta_2=0.99$. 
The deep feature extraction module of the proposed network consists of six MDDBs, each containing six MDDLs. The number of hidden feature dimensions is 156. The experiments are conducted using the PyTorch framework on an NVIDIA GeForce RTX 4090 GPU.

\subsection{Comparisons with State-of-the-Art Methods}
\subsubsection{Quantitative Results}
To evaluate the performance of the proposed method, comparisons are made with state-of-the-art (SOTA) 2D planar image SR methods and ODISR methods. The 2D planar image SR methods selected for comparison include SwinIR \cite{liang2021swinir}, HAT \cite{chen2023activating}, and RGT \cite{chen2024recursive}, while the ODISR methods include OSRT \cite{yu2023osrt}, BPOSR \cite{wang2024omnidirectional}, and GDGT-OSR \cite{yang2025geometric}. To ensure fair comparison, all methods are trained under identical conditions. The lightweight version of GDGT-OSR is used, which has a comparable number of parameters to the proposed model. Table~\ref{SOTA} shows the comparison results across the three test datasets, i.e., the ODI-SR, SUN 360 Panorama, and Flickr360 val datasets, under $\times{4}$, $\times{8}$ and $\times{16}$ scaling factors. Due to their inability to effectively transform geometric distortion, the 2D planar image SR methods, such as HAT \cite{chen2023activating} and RGT \cite{chen2024recursive}, which perform well on planar images, experience significant degradation when applied to ODIs, especially at large scaling factors. In contrast, the proposed MDDN outperforms all 2D planar image SR methods and ODISR models, demonstrating superior performance in handling omnidirectional images. Compared to GDGT-OSR \cite{yang2025geometric}, which is an ODISR method, MDDN achieves performance gains of 0.20 dB and 0.16 dB, in terms of PSNR and WS-PSNR, respectively, on the Flickr360 val dataset under the $\times{4}$ scaling factor. These results demonstrate that the proposed MDDN can effectively leverage multi-level feature representations, leading to remarkable performance in ODISR.

\subsubsection{Qualitative results}
\begin{figure*}
\centering
\includegraphics[width=\textwidth]{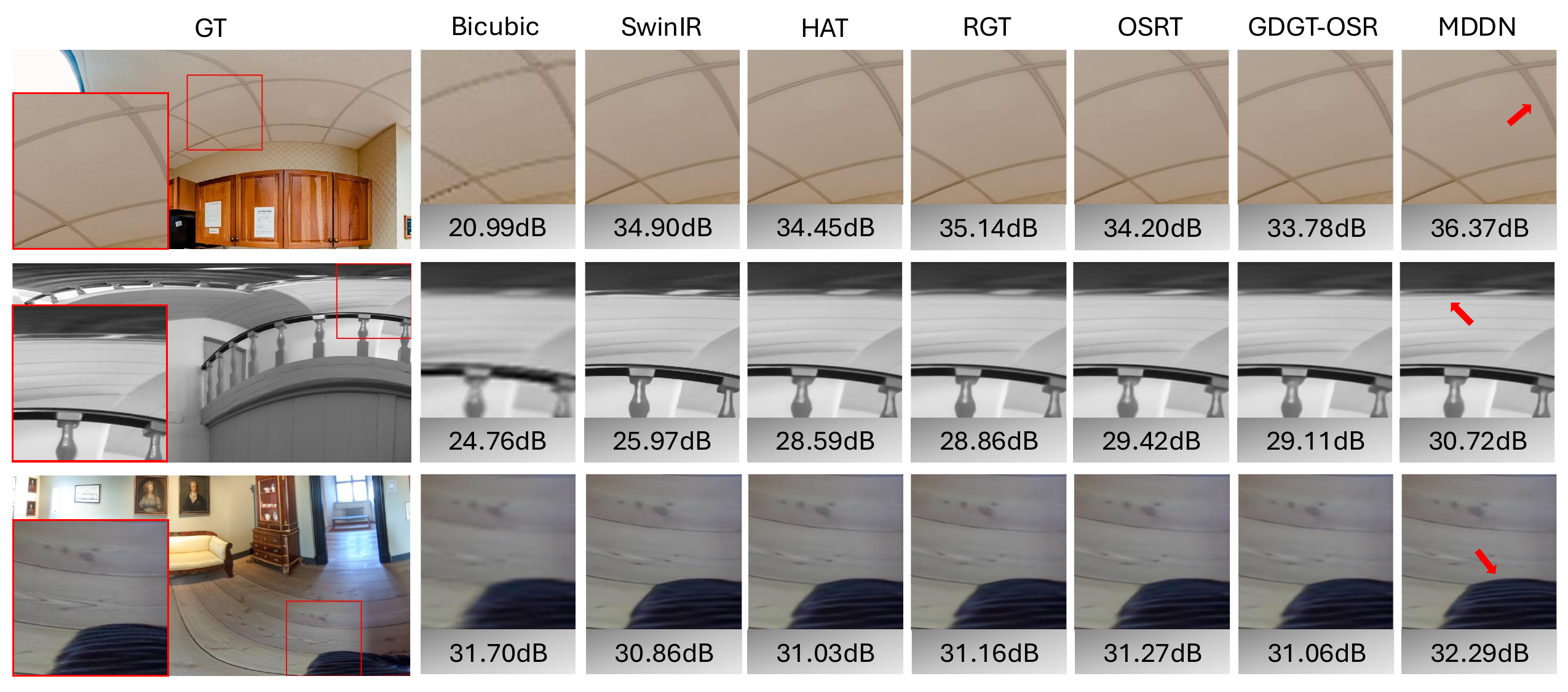}
\caption{Visual comparisons among different methods. The upscaling factor is 4. The PSNR values are calculated between the patches generated from the methods and the ground-truth patch, which are shown below the corresponding patches.}
\label{visual_x4}
\end{figure*}
The qualitative evaluation of different SR methods under $\times4$ scaling factor is shown in Fig.~\ref{visual_x4}. The PSNR is shown under the corresponding patch.
In the first sample, the ceiling in the high latitude is distorted significantly, and the repetitive pattern is sparse. Thanks to the strong distortion transformation ability and the large sampling range, the proposed MDDN can achieve the best performance in both quantitative metric and visual result among the compared methods.
In the second sample, the texture in the highlighted patch is more complicated. Both the 2D planar image SR methods and the ODISR methods struggle to restore the fine details and almost smooth the edges (red arrow), while the proposed method can restore the fine edges, which are closer to the ground truth. 
The third sample consists of the wooden floor and the striped shirt of the cameraman, which involves many intricate textures and is difficult to restore, especially the striped shirt. As is shown in Fig.~\ref{visual_x4}, the other SOTA SR methods, like SwinIR \cite{liang2021swinir}, HAT \cite{chen2023activating} and RGT \cite{chen2024recursive}, smooth some parts of the stripes. Even though OSRT \cite{yu2023osrt} and GDGT-OSR \cite{yang2025geometric} are specifically designed for handling ODIs, they still find it hard to restore the details of the stripes due to their limited receptive field and inadequate feature extraction capabilities for ODIs. Benefiting from the combination of global and local feature extraction, our proposed method can not only restore the ground well, but also preserve the fine textures of the stripes accurately, achieving a high PSNR value. Note that the PSNR of the patch generated by bicubic interpolation is higher than those generated by other SR methods, except for MDDN. Nevertheless, though with high quantitative performance, the visual result of the patch generated by bicubic interpolation is the smoothest, lacking almost all of the details. Some oversmooth results may achieve good performance on pixel-level metrics, e.g., PSNR and SSIM, which is also an inherent drawback of these metrics.

\begin{table}[]
\centering
\caption{Results of variants with different combinations of feature extractors. The scaling factor is 4. The performance is evaluated on the Flickr360 val dataset.}
\scalebox{0.79}{
\begin{tabular}{c|cccc|c}
\hline
Variants    & PSNR           & SSIM            & WS-PSNR        & WS-SSIM         & \#Params(M) \\ \hline\hline
\{1\}       & 30.25          & 0.8337          & 29.72          & 0.8259          &  11,73       \\ \hline
\{2\}       & 30.10          & 0.8297          & 29.57          & 0.8219          &  7.82       \\ \hline
\{3\}       & 30.10          & 0.8296          & 29.57          & 0.8219          &  7.82       \\ \hline
\{1, 2\}    & 30.27          & 0.8339          & 29.73          & 0.8259          &  12.41       \\ \hline
\{1, 3\}    & 30.27          & 0.8338          & 29.74          & 0.8260          &  12.41       \\ \hline
\{2, 3\}    & 30.21          & 0.8327          & 29.65          & 0.8241          &  8.50       \\ \hline
\{1, 2, 3\} & 30.44          & 0.8380          & 29.87          & 0.8295          &  13.04      \\ \hline
\{1, 2, 4\} & 30.44          & 0.8381          & 29.87          & 0.8296          &  13.04       \\ \hline
\{1, 3, 4\} & 30.44          & 0.8379          & 29.86          & 0.8294          &  13.04       \\ \hline
\{1, 2, 5\}  & 30.44          & 0.8380          & 29.87          & 0.8295          &  13.04        \\ \hline
\{1, 2, 3, 4\} & 30.45        & 0.8382          & 29.88          & 0.8297          &  13.67        \\ \hline
\end{tabular}
}
\label{variant}
\end{table}

\begin{table*}[]
\centering
\caption{Ablation study of low-rank decomposition.}
\begin{tabular}{c|cccc|cc}
\hline
  & PSNR  & SSIM   & WS-PSNR & WS-SSIM & \#Params(M) & Multi-Adds(G) \\ \hline\hline
Less Dim & 27.32 & 0.7724 & 26.61 & 0.7569  & 13.05 & 45.34       \\ \hline
Full Dim & \textbf{27.40} & \textbf{0.7751} & \textbf{26.69} & \textbf{0.7597}  & 30.62 & 56.50        \\ \hline
MDDN  & 27.39 & \textbf{0.7751} & 26.68 & 0.7595  & 13.04 & 45.41  \\ \hline
\end{tabular}
\label{decompose}
\end{table*}

\begin{figure*}
\centering
\includegraphics[width=\textwidth]{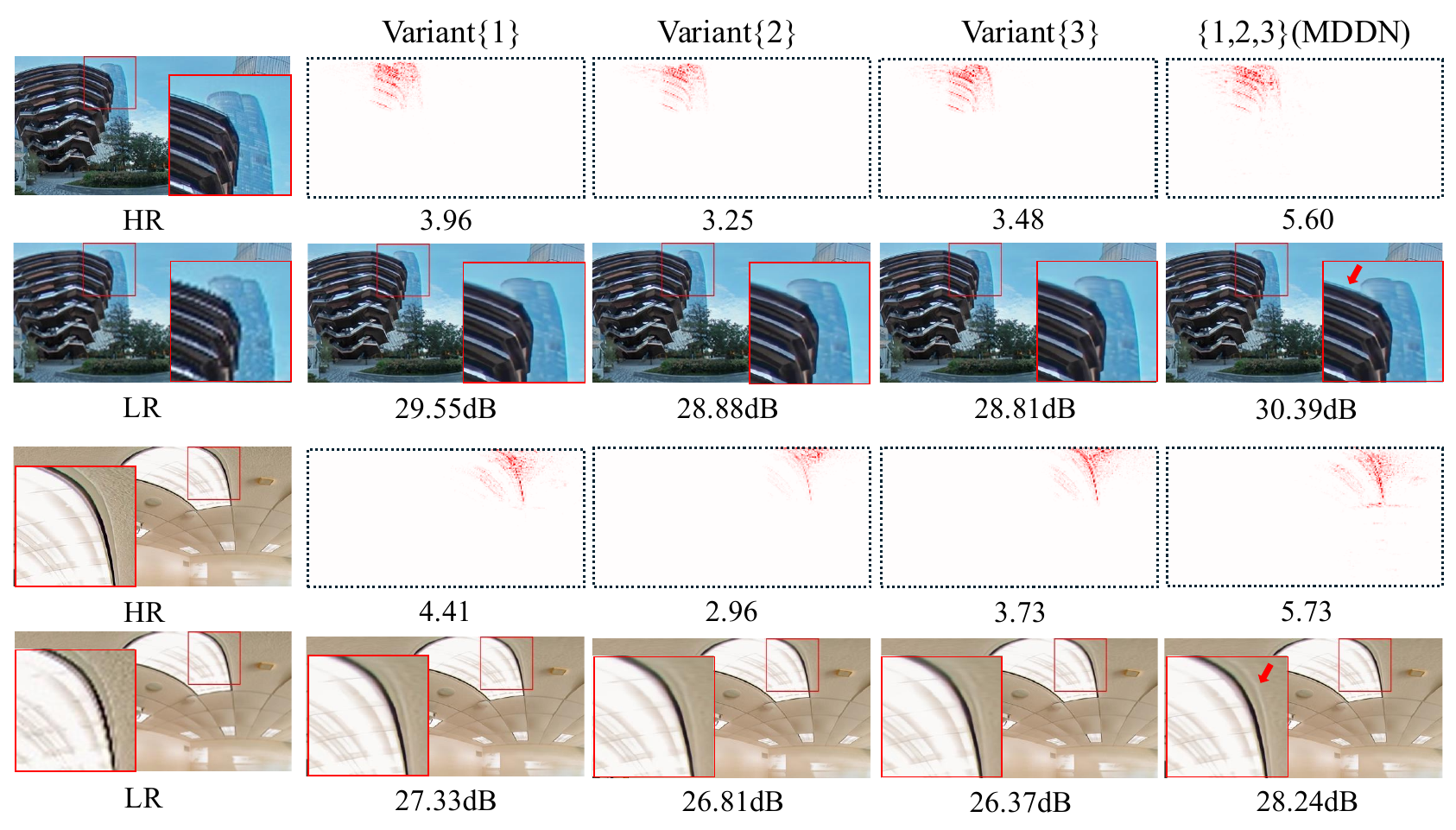}
\caption{Comparison of LAMs and visual results between the proposed MDDN and other single-level variants. The diffusion index (DI) is shown below each LAM, where a higher DI value corresponds to a wider range of pixels involved in the reconstruction of the patch highlighted by the red box.}
\label{LAM_variant}
\end{figure*}

\subsection{Ablation studies}
\subsubsection{Combination of the Multi-level Extractor}
\label{abla_mdde}
Table~\ref{variant} presents a quantitative comparison of variants with different combinations of feature extractors. 
We use the notation \{$1, ..., n$\} to represent variants with feature extractors of different levels. For example, the variant \{1, 2\} uses level-1 $(d=1)$ and level-2 $(d=2)$ extractors in the MDDE, while \{1, 2, 5\} consists of level-1 $(d=1)$, level-2 $(d=2)$, and level-5 $(d=5)$ extractors. 
From Table~\ref{variant}, we observe that the variants with single-level feature extractors, i.e., \{1\}, \{2\}, and \{3\}, perform significantly worse than multi-level variants. Among them, variant \{1\} outperforms \{2\} and \{3\}, highlighting the fundamental importance of global representation provided by the level-1 feature extractor. 
To further visualize the difference between MDDN and single-level variants, we employ the local attribution map (LAM), which is an attribution tool designed to visualize pixel importance in SR reconstruction.
Fig.~\ref{LAM_variant} compares LAMs and visual results between our MDDN and single-level variants. LAMs highlight the importance of specific pixels in the reconstruction of the patch enclosed in the red box. The diffusion index (DI) displayed below each LAM quantifies the range of involved pixels, where higher DI values mean wider pixel coverage. As shown in Fig.~\ref{LAM_variant}, compared to single-level variants, the LAM of the proposed MDDN demonstrates a larger sampling range and more involved pixels, reflected by a higher DI value. Visual results also show that edges in the outputs of MDDN are sharper and clearer than those outputs from single-level variants. This suggests that a larger sampling range of related and self-similar pixels can contribute to improved SR reconstruction. The global and local feature extractors in MDDN equip the model with a larger sampling range, enhancing reconstruction quality. Furthermore, adding an extra feature extractor consistently improves the performance of single-level variants, as seen in variants \{1, 2\}, \{1, 3\}, and \{2, 3\} in Table~\ref{variant}, demonstrating the cooperation and complementarity between feature extractors. 
The model incorporating all three types of extractors, i.e., level-1, level-2, and level-3 extractors, achieves a significant performance boost. This shows that global and local feature extractors complement each other effectively, and each plays a crucial role.
We further investigate the impact of higher-level feature extractors, such as level-4 and level-5, in combinations like \{1, 2, 4\}, \{1, 3, 4\}, and \{1, 2, 5\}. Although these variants include higher-level extractors, with larger sampling ranges, their performance is comparable to, or even worse than, \{1, 2, 3\}. This suggests that excessively large sampling ranges may incorporate irrelevant pixels, leading to deteriorating reconstruction quality easily.
Similarly, adding a fourth extractor, e.g., \{1, 2, 3, 4\}, increases parameters without delivering a clear performance gain. 
Based on this analysis, we adopt the variant \{1, 2, 3\} as the default configuration for multi-level feature extractors in our model.

\begin{table*}[]
\centering
\caption{Ablation study of the rank $(r)$. The scaling factor is 4. The evaluation is conducted on the ODI-SR dataset.}
\begin{tabular}{c|cccc|cc}
\hline
$r$  & PSNR  & SSIM   & WS-PSNR & WS-SSIM & \#Params(M) & Multi-Adds(G) \\ \hline\hline
4  & 27.32 & 0.7721 & 26.61   & 0.7569  & 12.47 & 44.90      \\ \hline
8  & 27.39 & 0.7751 & \textbf{26.68}   & 0.7595  & 13.04 & 45.41      \\ \hline
12 & \textbf{27.40} & \textbf{0.7752} & \textbf{26.68}   & \textbf{0.7596}  & 13.62 & 45.92  \\ \hline
16 & \textbf{27.40} & \textbf{0.7752} & \textbf{26.68}   & \textbf{0.7596}  & 14.19 & 46.43  \\ \hline
20 & \textbf{27.40} & \textbf{0.7752} & \textbf{26.68}   & \textbf{0.7596}  & 14.64 & 46.94  \\ \hline
\end{tabular}
\label{rank}
\end{table*}

\begin{table}[]
\centering
\caption{Results of different approaches in fusing outputs from branches. The scaling factor is 4. The performance is evaluated on the Flicker360 val dataset.}
\begin{tabular}{c|cccc}
\hline
      & PSNR  & SSIM   & WS-PSNR & WS-SSIM \\ \hline\hline
Addition & 30.38 & 0.8364 & 29.82   & 0.8283  \\ \hline
MFF   & \textbf{30.44} & \textbf{0.8380} & \textbf{29.87}   & \textbf{0.8295}  \\ \hline
\end{tabular}
\label{fuse}
\end{table}

\subsubsection{Impacts of Low-Rank Decomposition} 
To reduce computational cost, we adopt low-rank decomposition to factorize a large kernel weight matrix into two low-rank matrices, improving model efficiency. To evaluate its effectiveness, we conducted experiments comparing MDDN with variants that do not apply the low-rank decomposition strategy. 
The first variant, Less Dim, forgoes low-rank decomposition and instead directly reduces channel dimensionality in the offset network and D4C layers, matching its total parameter count to MDDN. As shown in Table~\ref{decompose}, despite having a similar number of parameters, the Less Dim variant exhibits inferior performance. This demonstrates that low-rank decomposition extracts more effective representations than simple dimensionality reduction, highlighting the superiority of our low-rank decomposition strategy.
% not only reduces model complexity but also improves efficiency and accuracy. 
We further introduce a full-dimensional variant, Full Dim, which retains the original feature dimensionality $C_1$ without low-rank decomposition.
Compared to MDDN, this variant has more than twice as many parameters and substantially higher computational complexity.
However, it achieves only comparable performance, with no clear gain. This indicates considerable redundancy in the full-dimensional feature maps.
% Fig.~\ref{decomposition} shows the PSNR performance of MDDN with $r=8$ and the `full dim' variant on the test set, measured every 100 iterations during the training process. 
% As shown in Fig.~\ref{decomposition}, MDDN's performance improves steadily with increasing iterations. In contrast, despite using full weights in the D4C layers and offset networks, the PSNR performance of `full dim' drops sharply after 80,000 iterations, confirming that excessive trainable parameters lead to severe overfitting. 
These experimental results demonstrate that adopting low-rank decomposition is both necessary and beneficial for improving model performance while reducing complexity.

\subsubsection{Impacts of Rank}
We explore the effect of different rank values in the low-rank decomposition of the D4C layer and offset networks. As shown in Table~\ref{rank}, performance improves as the rank increases from 4 to 8. 
% However, when the rank is further increased (beyond 8 to 10), performance slightly declines even though the number of parameters increases. 
When the rank increases further from 8 to 20, the model's performance gain becomes marginal, while the number of parameters and model complexity increase significantly, leading to more training memory and computational cost.
% However, the model deteriorates with an even larger rank, such as $r=20$, which suggests that an excessive rank may result in model overfitting due to the increased number of training parameters. 
% The extreme case, shown as `full dim' in Table~\ref{decompose} and Fig.~\ref{decomposition}, indicates that an excessive number of parameters not only aggravates model complexity but also easily causes model overfitting.
% When the rank increases to the extreme case, shown as `full dim' in Table~\ref{decompose}, which does not apply low-rank decomposition to the layers, the number of parameters and computational complexity have grown dramatically, yet the model's performance shows no obvious improvement.
In the extreme case, shown as `Full Dim' in Table~\ref{decompose}, where low-rank decomposition is entirely disabled, the model exhibits a drastic increase in both parameter count and computational complexity, with no significant gain in performance.
Therefore, to balance the model complexity and the performance, we adopt $r=8$ as the default rank for our experiments.

\subsubsection{Impacts of Multi-level Feature Fusion}
The three output features from the multi-level extractors are adaptively aggregated in the MFF module. 
% To evaluate different fusion strategies, we compare several fusion approaches in Table~\ref{fuse}. 
We compare different fusion approaches in Table~\ref{fuse}. 
The `Addition' approach represents that the three features from the extractors are fused through direct pixel-wise addition. As shown in Table~\ref{fuse}, performance decreases across all evaluation metrics when features are fused through direct addition. This decline may be attributed to the geometric properties of ERP images, where distortion varies by latitude. Different regions require distinct feature extraction ranges, with highly distorted areas necessitating a larger extraction range due to extreme scene expansion. The model can benefit from an adaptive feature fusion strategy, which optimally aggregates feature representations across varying distortion levels.

\begin{figure*}
\centering
\includegraphics[width=\textwidth]{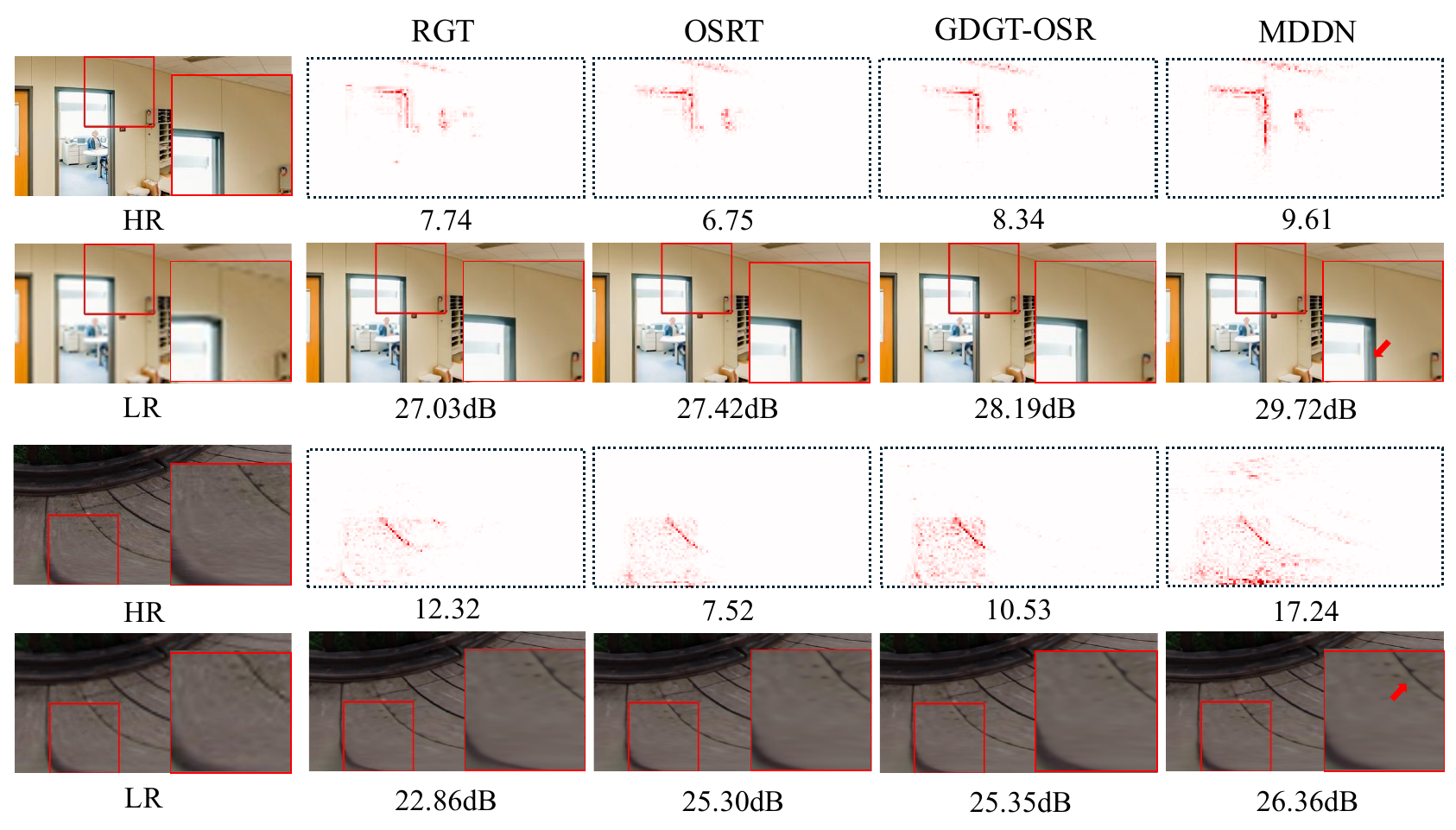}
\caption{Comparison of local attribution maps (LAMs) and visual results of different methods. The diffusion index (DI) is shown below each LAM, where a higher DI value corresponds to a wider range of pixels involved in the reconstruction of the patch in the red box.}
\label{LAM}
\end{figure*}

\subsection{Exploration of the Involved Area}
To analyze the range of pixels involved in other SR methods compared to the proposed method, we examine their LAMs \cite{gu2021interpreting} and visual results in Fig.~\ref{LAM}.
% we employ the local attribution map (LAM) \cite{gu2021interpreting}, which is an attribution tool designed for visualizing pixel importance in SR reconstruction. Fig.~\ref{LAM} visually illustrates the LAM results and SR results for different methods. 
LAMs highlight the importance of corresponding pixels in reconstructing the patch. We also present the DI value below each LAM, which quantifies the extent of pixel involvement. A higher DI value means wider pixel coverage. 
As shown in the two samples in Fig.~\ref{LAM}, the proposed method achieves the widest contribution area and the highest DI values. This indicates a large sampling range, enabling the capture of more self-similar textures due to the multi-level feature extractor design, which is important for accurate ODISR reconstruction. For example, in the second sample, the contribution areas in LAMs of other methods, such as RGT \cite{chen2024recursive}, OSRT \cite{yu2023osrt}, and GDGT-OSR \cite{yang2025geometric}, are almost confined within the red box. This limitation reduces the ability to locate more similar regions for improved reconstruction. In contrast, the proposed method, thanks to its multi-level feature extractors that combine global and local feature extraction, extends the contribution area beyond the red box to include more similar regions.
By incorporating more pixels from self-similar textures, the proposed MDDN delivers the most visually appealing SR results and achieves highest PSNR.

\begin{table}[]
\centering
\caption{Model complexity comparisons among different methods.}
% \resizebox{0.49\textwidth}{!}{
\scalebox{0.77}{
\begin{tabular}{c|c|c|c|c}
\hline
Model    & \#Params(M) & \#Multi-Adds(G) & PSNR  & WS-PSNR \\ \hline\hline
SwinIR \cite{liang2021swinir}   & 12.05       & 60.59           & 26.79 & 26.25   \\ \hline
HAT \cite{chen2023activating, chen2023hat}     & 20.92       & 96.05           & 26.57 & 26.00   \\ \hline
RGT \cite{chen2024recursive}     & 13.51       & 62.17           & 26.79 & 26.25   \\ \hline
OSRT \cite{yu2023osrt}    & 12.08       & 54.50           & 26.94 & 26.35   \\ \hline
BPOSR \cite{wang2024omnidirectional}   & 2.19        & 40.33           & 25.26 & 24.74   \\ \hline
GDGT-OSR \cite{yang2025geometric} & 13.45       & 56.69           & 26.94 & 26.36   \\ \hline
MDDN     & 13.19       & 55.43           & 27.05 & 26.45   \\ \hline
\end{tabular}
}
\label{complexity}
\end{table}

\subsection{Model Complexity}
Table~\ref{complexity} presents the model complexity of different methods in terms of the number of parameters and Multi-Adds. Performance is evaluated on the Flickr360 val dataset under the $\times8$ scaling factor, with Multiply-Adds calculated for an input size of $64\times64$ pixels. Although 2D planar image SR methods, such as HAT \cite{chen2023activating} and RGT \cite{chen2024recursive}, have a higher number of parameters and Multiply-Adds compared to ODISR methods, such as OSRT \cite{yu2023osrt}, GDGT-OSR \cite{yang2025geometric}, and our proposed MDDN, their performance remains inferior to ODISR-specific methods. This also illustrates that tailored ODISR methods are needed to process ODIs. The number of parameters and Multiply-Adds of the lightweight version of GDGT-OSR \cite{yang2025geometric} is larger than those of our MDDN. However, with less model complexity, MDDN still surpasses the SOTA GDGT-OSR. This demonstrates the effectiveness and superiority of the proposed MDDN in terms of model complexity. 

% \section{Summary}
\subsection{Limitations}
% \subsubsection{Lack of Semantic Information}
Our method leverages the Transformer architecture as the backbone, which excels in capturing long-range dependencies and modeling complex patterns within data. However, one notable limitation is its lack of explicit semantic information utilization. The Transformer model primarily focuses on low-level and mid-level feature extraction, often overlooking higher-level semantic context, which could enhance its understanding of the input. Recent diffusion-based SR models emphasize the importance of semantic information in improving reconstruction performance. These models effectively integrate semantic cues, such as class labels and text descriptions, to guide the process. In contrast, our method only relies on image-based features, which limits its flexibility and expressiveness in scenarios where semantic guidance is critical. 

\section{Conclusion}
This paper introduces the Multi-level Distortion-aware Deformable Network (MDDN), a novel approach for effective multi-level representation extraction in omnidirectional image super-resolution (ODISR). By integrating a distortion-aware deformable attention mechanism with dilated distortion-aware deformable convolutions, MDDN is equipped with diverse feature extractors with multi-level sampling ranges that adapt to distorted ERP images, utilizing complementary global and local features. These multi-level feature extractors cooperate adaptively in the multi-level feature fusion module, enhancing reconstruction performance. In addition, a low-rank decomposition strategy is adopted to reduce computational cost and optimize efficiency. Extensive experiments on public datasets demonstrate that MDDN outperforms both 2D planar super-resolution and existing ODISR methods, achieving superior quantitative and qualitative results, thereby confirming the effectiveness and superiority of the proposed method.

\bibliographystyle{cas-model2-names}

\bibliography{refs}

\end{document}